\documentclass[sigconf]{acmart}
\AtBeginDocument{%
  }

\setcopyright{acmlicensed}
\copyrightyear{2018}
\acmYear{2018}
\acmDOI{XXXXXXX.XXXXXXX}
\acmConference[Conference acronym 'XX]{Make sure to enter the correct
  conference title from your rights confirmation email}{June 03--05,
  2018}{Woodstock, NY}
\acmISBN{978-1-4503-XXXX-X/2018/06}

\usepackage{CJKutf8}
\usepackage{graphicx}
\usepackage{multirow}

\usepackage[linesnumbered,ruled,vlined]{algorithm2e}

\SetKwInput{KwInput}{Input}                
\SetKwInput{KwOutput}{Output}              

\begin{document}

\title{Learning Novel Transformer Architecture for Time-series Forecasting}

\author{Juyuan Zhang}
\affiliation{%
  \institution{Nanyang Technological University}
  \city{SG}
  \country{Singapore}
  }
\email{JZHANG161@e.ntu.edu.sg}

\author{Wei Zhu}
\affiliation{%
  \institution{University of Hong Kong}
  \city{Hong Kong}
  \country{China}}
\email{michaelwzhu91@gmail.com}

\author{Jiechao Gao}
\affiliation{%
  \institution{University of Virginia}
  \city{VA}
  \country{United States}}
\email{jg5ycn@virginia.edu}

\renewcommand{\shortauthors}{Trovato et al.}

\begin{abstract}

Despite the success of Transformer-based models in the time-series prediction (TSP) tasks, the existing Transformer architecture still face limitations and the literature lacks comprehensive explorations into alternative architectures. To address these challenges, we propose AutoFormer-TS, a novel framework that leverages a comprehensive search space for Transformer architectures tailored to TSP tasks. Our framework introduces a differentiable neural architecture search (DNAS) method, AB-DARTS, which improves upon existing DNAS approaches by enhancing the identification of optimal operations within the architecture. AutoFormer-TS systematically explores alternative attention mechanisms, activation functions, and encoding operations, moving beyond the traditional Transformer design. Extensive experiments demonstrate that AutoFormer-TS consistently outperforms state-of-the-art baselines across various TSP benchmarks, achieving superior forecasting accuracy while maintaining reasonable training efficiency. \footnote{Code will be publicly available upon acceptance. }

\end{abstract}



\received{20 February 2007}
\received[revised]{12 March 2009}
\received[accepted]{5 June 2009}

\maketitle

\begin{CJK*}{UTF8}{gbsn}

\section{Introduction}
\label{sec:intro}

Time series forecasting (TSP) represents a crucial modeling endeavor \cite{jin2023large}, spanning a wide array of practical applications such as climate modeling, inventory management, and energy demand prediction. Many efforts have been devoted in proposing different types of models to enhance the performance of TSP. In the 1970s, the dominant approaches are the statistical models like ARMA, ARIMA \cite{box2015time}, GARCH \cite{bauwens2006multivariate}, and structural models \cite{bollen1989structural}. With the raise of machine learning (ML) in the 1990s, many ML approaches have been applied to TSP, such as SVM \cite{meyer2001support}, decision tree, and ensemble models like Gradient boosting regression tree (GBRT) \cite{drucker1994boosting}. In the deep learning (DL) era, various deep learning-based time series models, including recurrent neural networks, convolutional neural networks, linear models and Transformer \cite{wen2022transformers}, are proposed in the literature. Among these models, Transformer \cite{vaswani2017attention} now plays an important role in the recent works like iTransformer \cite{liu2023itransformer} and PatchTST \cite{nie2022time}, achieving better performances across different benchmark tasks.

With the widespread applications of Transformer in different domains and modalities, the research field has witnessed many sophisticated Transformer variants tailored for TSP \cite{zhou2021informer,kitaev2020reformer,wu2021autoformer,zhou2022fedformer}. However, the existing methods have the following limitations. First, a branch of Transformer-based methods focused on manually altering the attention mechanisms \cite{zhou2021informer}. However, these works focused on designing more efficient attention functions, reducing the overall complexity. However, as \cite{nie2022time} demonstrates, these models fail to perform well on the TSP tasks, and have been outperformed by simple architectures like multi-layer perceptrons (MLPs). Second, recent models like iTransformer \cite{liu2023itransformer}  and PatchTST \cite{nie2022time} have shown that with the proper tokenization methods, the Transformer model can achieve the state-of-the-art (SOTA) performance with the original architecture design. However, alternative architectures have not been fully investigated.

\begin{figure*}[tb!]
\centering
\includegraphics[width=0.7\textwidth]{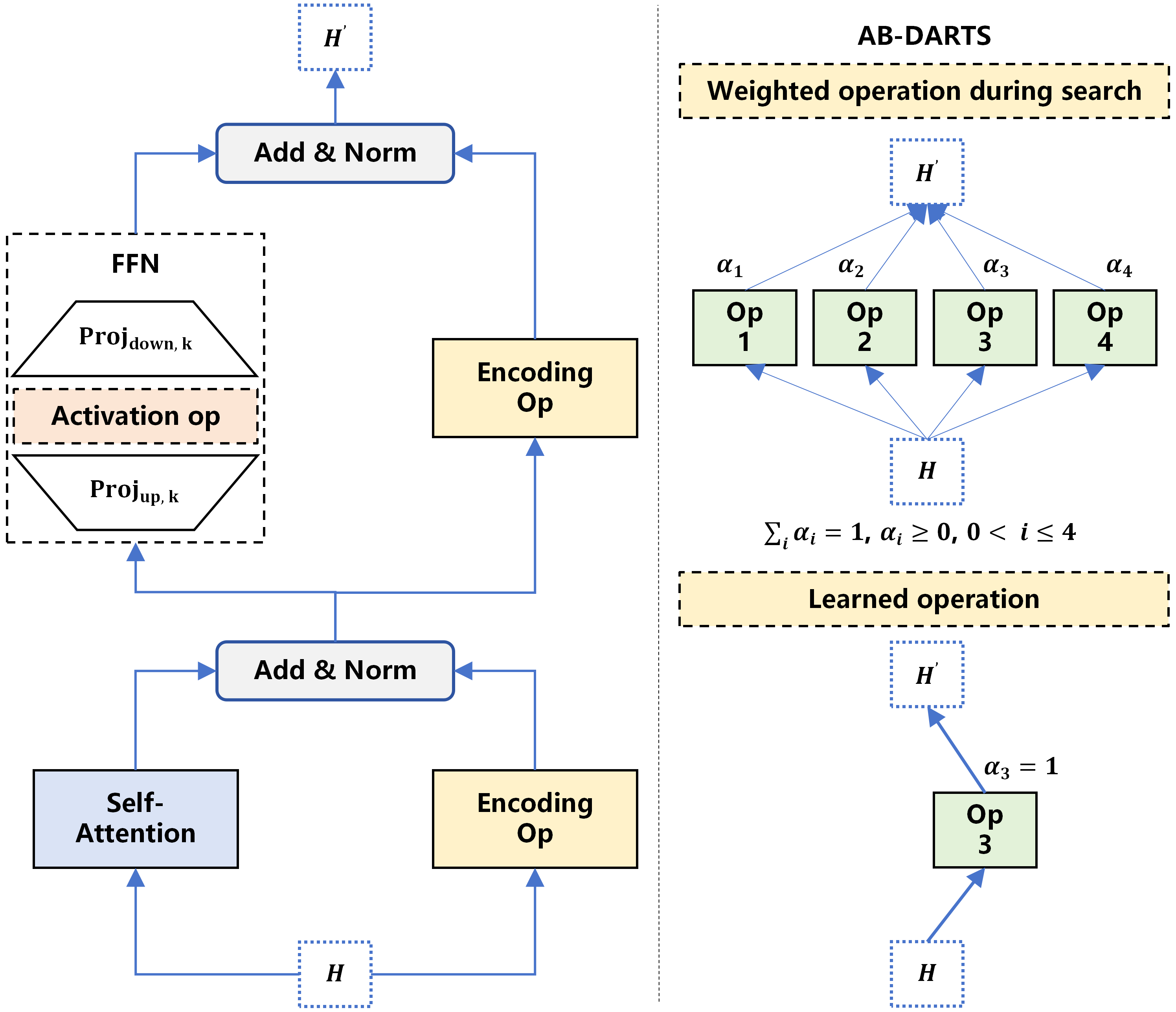}
\caption{The architecture for our AutoFormer-TS framework
. }
\label{fig:meta_arch}
\end{figure*}

To address the above issues, we propose the AutoFormer-TS framework (see Figure \ref{fig:meta_arch}). First, we look into the architecture of the vanilla Transformer, and design a comprehensive search space for TSP Transformer models. Our search space contains alternative attention mechanisms, activation functions, and encoding operations that substitute the residual connections. As for the search algorithm, we employ the differentiable neural architecture search (DNAS) \cite{liu2018darts} framework. We have looked into the workflow of DNAS and find that the current DNAS methods may have shortcomings in identifying which operations to keep. Thus, we propose a novel DNAS method, AB-DARTS. 

Extensive experimentation has proved that our AutoFormer-TS framework can successfully identify architectures that surpasses recent state-of-the-art baseline methods on the TSP task at hand. The contributions of our work are summarized as follows:
\begin{itemize}
\item Our AutoFormer-TS framework constructs a comprehensive and compact search space, which identifies the alternative operations or functions that could benefit the Transformer architecture. 

\item Our AutoFormer-TS framework proposes a novel DNAS method, AB-DARTS, which modifies the mechanism of identifying the most contributing operation on an edge of the hyper-network.

\item AutoFormer-TS consistently exceeds the state-of-the-art performance in TS forecasting tasks, while requiring reasonable training time cost. 

\end{itemize}

\section{Related works}
\label{sec:related_works}

\subsection{Time series modeling }

As a classical research problem with widespread applications, models constructed from the statistical approaches for time series modeling have been used from the 1970s. The representative models are autoregressive integrated moving average (ARIMA) \cite{box2015time}, exponential smoothing \cite{gardner1985exponential}, and structural models \cite{bollen1989structural}. The most significant characteristic for these methods is that they require significant domain expertise to build. With the development of machine learning (ML) \cite{biship2007pattern}, many ML techniques are introduced to time series modeling to reduce manual efforts. Gradient boosting regression tree (GBRT) \cite{drucker1994boosting,prokhorenkova2018catboost} gains popularity by learning the temporal dynamics of time series in a data-driven manner. However, these methods still require manual feature engineering and model designs. With the powerful representation learning capability of deep learning (DL) from large-scale data, various deep learning-based time series models are proposed in the literature \cite{lim2021time}, achieving better forecasting accuracy than traditional techniques in many cases. Before the era of Transformer \cite{vaswani2017attention}, the two popular DL architectures are: (a) Recurrent neural networks (RNNs) based methods \cite{Hochreiter1997LongSM}, which summarize the past information compactly in internal memory states and recursively update themselves for forecasting. (b) Convolutional neural networks (CNNs) based methods \cite{li2021survey}, wherein convolutional filters are used to capture local temporal features. More recently, multi-layer perceptron (MLP) based methods, like \cite{tang2025ts} and \cite{zeng2023transformers}, have raised attention in the research field, since these models are simple and light-weight.

\subsection{Transformer architectures in time series}

The progressive advancements in natural language processing and computer vision have led to the development of sophisticated Transformer \cite{Vaswani2017AttentionIA} variants tailored for a wide array of time series forecasting applications \cite{zhou2021informer,wu2021autoformer}. Central to these innovations is the methodology by which Transformers handle time series data. For instance, iTransformer \cite{liu2023itransformer} treats each univariate time series as a distinct token, forming multivariate time series into sequences of such tokens. More recently, PatchTST \cite{nie2022time} adopts an assumption of channel independence, transforming a univariate time series into multiple patches, which are subsequently treated as tokens and processed through a Transformer encoder. Another important research direction is to design alternative Transformer architectures. This branch of works mainly devote themselves into manually designing novel attention mechanisms, including  Reformer \cite{kitaev2020reformer}, Informer \cite{zhou2021informer}, AutoFormer \cite{wu2021autoformer}, FEDformer\cite{zhou2022fedformer}.

\subsection{Neural architecture search methods}

In the early attempts, NAS requires massive computations, like thousands of GPU days~\cite{Zoph2017NeuralAS,Zoph2018LearningTA,Liu2018ProgressiveNA}. Recently, a particular group of one-shot NAS, led by the seminal work DARTS~\cite{Liu2019DARTSDA} has attracted much attention. DARTS formulates the search space into a super-network that can adjust itself in a continuous space so that the network and architectural parameters can be optimized alternately (bi-level optimization) using gradient descent. A series of literature try to improve the performance and efficiency of DARTS, such as \cite{Xie2019SNASSN,Chen2021ProgressiveDB,Chu2021FairNASRE,Nayman2019XNASNA}. SNAS~\cite{Xie2019SNASSN} reformulate DARTS as a credit assignment task while maintaining the differentiability. \cite{Gao2020MTLNASTN} penalize the entropy of the architecture parameters to encourage discretization on the hyper-network. P-DARTS~\cite{Chen2021ProgressiveDB} analyze the issues during the DARTS bi-level optimization, and propose a series of modifications. PC-DARTS~\cite{Xu2021PartiallyConnectedNA} reduces the memory cost during search by sampling a portion of the channels in super-networks. FairDARTS~\cite{Chu2021FairNASRE} change the softmax operations in DARTS into sigmoid and introduce a zero-one loss to prune the architectural parameters. XNAS~\cite{Nayman2019XNASNA} dynamically wipes out inferior architectures and enhances superior ones.

Our work complements the literature by the following two aspects: (a) we conduct a pilot experiment to analyze the shortcomings of the current DNAS methods; (b) we propose a novel DNAS method that can achieve better search performances and search stability.

\section{Search Space Design}
\label{sec:search_pace_design}

Now we discuss our search space in detail. Since our goal here is to optimize the transformer architecture, we keep its main bone structure, as shown in Figure~\ref{fig:meta_arch}. We design a comprehensive search space for discovering discovering novel
self-attention structures as well as the novel architectures for the pointwise feed-forward module.

\subsection{Search Space for the self-attention module}

As shown in Figure \ref{fig:meta_arch}, the original self-attention mechanism in the Transformer model can be expressed as follows:
\begin{align}
q_i & = x_{i}W_{Q}, \nonumber \\
k_j & = x_{j}W_K, \nonumber \\
s_{i, j} & = q_{i}^{\top} \odot k_j, \nonumber \\
\label{eq:dot_attn}
\end{align}
where $ x_{i}, x_{j} \in \mathbb{R}^{l \times d_{m}} $ is the input tokens $i$ and $j$'s hidden representations. $i, j \leq l$, where $l$ is the sequence length and $d_{m}$ is the model's hidden dimension. The above dot-product attention mechanism (denoted as Dot\_Attn), i.e., how to compute the attention scores $s_{i, j}$ between any token pairs, is the core to the self-attention module \cite{vaswani2017attention}.\footnote{Here we do not include the multi-head aspect in the attention module to simplify the math expressions. However, we will implement the multi-head attention module in the experiments. } 

Although Dot\_Attn in Equation \ref{eq:dot_attn} is the most widely used one in the Transformer models in the literature. However, there are also a wide collection of attention functions in the literature. Here we present four alternative attention mechanisms:
\begin{itemize}
\item Elementwise product attention \cite{rocktaschel2015reasoning} (denoted as EP\_Attn), where the attention score is calculated as the Hardmard product between $q_{i}$ and $k_j$:
\begin{equation}
s_{i, j} = \text{tanh}(q_{i} \odot k_j )^{\top} W_{d},
\end{equation}
where $W_{d} \in \mathbb{R}^{d_{m} \times 1 }$ is the learnable parameter matrix, $\odot$ is the Hardmard product operation, and $\text{tanh}()$ is the hyperbolic tangent function. 

\item Bilinear attention \cite{wang2015learning} (Bilinear\_Attn), where the attention score is calculated as the bilinear product between $q_{i}$ and $k_j$:
\begin{equation}
s_{i, j} = q_{i}^{\top} W_{d} k_j,
\end{equation}
where $W_{d} \in \mathbb{R}^{d_{m} \times d_{m} }$ is the learnable parameter matrix.

\item Concat attention \cite{velivckovic2017graph} (Concat\_Attn), where the attention score is calculated by concatenating the query and key vectors and going through an activation function and a linear projection layer:
\begin{equation}
s_{i, j} = \text{tanh}( \text{concat}([q_{i}, k_j]) )^{\top} W_{d},
\end{equation}
where $W_{d} \in \mathbb{R}^{d_{m} \times 1 }$ is the learnable parameter matrix.

\item Minus attention \cite{tan2018multiway} (Minus\_Attn), where the attention score is calculated by letting the query vector to subtract the key vector and going through an activation function and a linear projection layer:
\begin{equation}
s_{i, j} = \text{tanh}( q_{i} - k_j )^{\top} W_{d},
\end{equation}
where $W_{d} \in \mathbb{R}^{d_{m} \times 1 }$ is the learnable parameter matrix.

\end{itemize}

In our search space, the above mentioned five attention functions will constitute the attention function search space.

\subsection{Search Space for the feed-forward module}

As shown in Figure \ref{fig:meta_arch}, the original feed-forward network (FFN) module is:
\begin{equation}
FFN(X) = \text{Proj}_{down, k} (g(\text{Proj}_{up, k} (X) ) ),
\end{equation}
where $\text{Proj}_{up, k}$ is the linear layer that projects the input from dimension $d_{m}$ to $d_{ffn}$. Here, $d_{ffn}$ denotes the intermediate dimension, $k$ denotes the dimension multiplication factor:
\begin{equation}
    d_{ffn} = k * d_{m}.
\end{equation}
$\text{Proj}_{down, k}$ projects the input from $d_{ffn}$ to $d_{m}$, and $g()$ is the activation function. In the original Transformer \cite{vaswani2017attention}, $k = 4$, and $g()$ is the ReLU activation. To construct the search space for the AutoFormer-TS framework, we now consider the following alternative design choices: (a) setting $k$ to one of \{ 0.5, 1, 2, 4 \}. (b) Setting the activation function $g()$ to ReLU \cite{he2016deep}, Leaky\_ReLU \cite{xu2015empirical}, SWISH \cite{ramachandran2017searching}, GeLU \cite{hendrycks2016gaussian} and ELU \cite{clevert2015fast}. The activation functions are presented in Table \ref{tab:activation_functions}.

\begin{table}
\centering
\caption{\label{tab:activation_functions}Activation functions in the search space.}
\resizebox{0.36\textwidth}{!}{
\begin{tabular}{lcc}
\hline \bf Name  & \bf Function  \\ 
\hline
ReLU &  $\max(x, 0)$    \\
Leaky\_ReLU  &  $x$ if $x \geq 0$ else 1e-2 * $x$    \\ 
ELU   &  $x$ if $x \geq 0$ else $e^{x} - 1$  \\
SWISH   & $x * Sigmoid(x)$   \\
GeLU   & $0.5*x*(1 + erf(x / \sqrt{2}))$  \\ 
\hline
\end{tabular}}

\end{table}

\subsection{Search Space for encoding operations}

In the original Transformer model, the two core modules include residue connections:
\begin{align}
X^{(1)} & = \text{Attn}(X) + X, \nonumber\\
X^{'} & = \text{FFN}(X^{(1)}) + X^{(1)}. \nonumber\\
\label{eq:transformer_block_2}
\end{align}
However, as demonstrated in \cite{so2019evolved}, when substituting some of the residue connections with proper encoding operations like convolutions, the model performance could improve. This observation is intuitive: different encoding operations could provide semantic information from different perspective, enriching the hidden representations of the look-back window. So the above equations become:
\begin{align}
X^{(1)} & = \text{Attn}(X) + \text{Enc\_{A}}(X), \nonumber\\
X^{'} & = \text{FFN}(X^{(1)}) + \text{Enc\_{F}}(X^{(1)}), \nonumber\\
\label{eq:transformer_block_3}
\end{align}
where $\text{Enc\_{A}}()$ and $\text{Enc\_{F}}()$ are encoding operations. Thus, it is natural for us to consider the following search space for the encoding operations in our AutoFormer-TS framework: 
\begin{itemize}
\item Special zero operation, denoted as Null;
	
\item Skip connection, denoted as Skip;
	
\item $1$-d convolutions, with kernel size $k$, where $k = 1, 3, 5$, denoted as Conv\_$k$;

\end{itemize}

\subsection{Summary of the whole search space}

Based on the above analysis of the alternative design choices of Transformer for the time series model, we now  formally introduce our whole search space:

\begin{itemize}

\item The encoding operation accompanying the self-attention module: \{ Null, Skip, Conv\_1, Conv\_3, Conv\_5 \};

\item The encoding operation accompanying the FFN module: \{ Null, Skip, Conv\_1, Conv\_3, Conv\_5 \};

\item The attention function $\text{Attn\_Func}$ in the self-attention module: \{Dot\_Attn, EP\_Attn, Bilinear\_Attn, Concat\_Attn, Minus\_Attn\};

\item The activation function $g()$ in the FFN module: \{ReLU, ELU, SWISH, Leaky\_ReLU, GeLU \};

\item The dimension multiplication factor $k$: \{0.5, 1, 2, 4\}.

\end{itemize}

For a Transformer model with 3 blocks, our search space contains 1.56e+10 number combinations of possible transformer architectures, which is quite a large search space. Next, we will show how to navigate through this enormous search space and obtain architectures that are better than standard Transformer model efficiently.

\section{Search algorithm}
\label{sec:search_algorithm}

\subsection{Preliminaries on differentiable neural architecture search}
\label{subsec:preliminaries_on_darts}

Now we give a brief introduction to the representative differentiable neural architecture search algorithm, DARTS \cite{liu2018darts}. During the search stage, DARTS initialize a hyper-network which is a connected directed acyclic graph (DAG) with $N$ nodes. Each node $\text{Node}_{i}$ is referred to as a search cell, which combines a set of $n_{i} $ operations $\{o_{i, j}\}_{j=1}^{n_{i}}$ via a weighted sum. Denote the architectural parameter for $o_{i, j}$ as $\alpha_{i, j} \in \mathbb{R}$, then the $\text{Node}_{i}$ represents the following operation: 
\begin{equation}
\text{Node}_{i}(x) = \sum_{j = 1}^{n_{i}} w_{i, j} o_{i, j} (x),
\end{equation}
in which $\{ w_{i, j}\}_{j=1}^{n_{i}}$ are given by:
\begin{equation}
w_{i, j} = \dfrac{ \exp{ \left( \alpha_{i, j} \right) }} { \sum_{j = 1}^{n_{i}} \exp{ \left(\alpha_{i, j} \right) } }.
\end{equation}
This design makes the entire framework differentiable to both layer weights and architectural parameters $\alpha_{i, j}$ so that it is possible to perform architecture search in an end-to-end fashion. The standard optimization method is the bi-level optimization proposed in DARTS, which splits the training set $\mathcal{D}_{train}$ into two subsets $\mathcal{D}_1$ and $\mathcal{D}_2$, one for network parameter updates and the other for updating the architectural parameters. Both groups of parameters are updated via gradient descent. After the search process is completed, the discretization procedure extracts the final sub-network by selecting the operation on each node with the highest $\alpha_{i, j}$ score and dropping the low-scored operations. And the final network will train on the original train set with randomly initialized parameters.

\subsection{Motivation} 
\label{sec:DNAS_motivation}

Note that the DARTS workflow select the operation on each node based on the architectural parameter $\alpha_{i, j}$, which is based on the hypothesis that the architectural parameter $\alpha_{i, j}$ can reliably reflect the quality or importance of operation $o_{i, j}$. However, We have conducted a pilot experiment demonstrating that architectural parameters in DARTS \cite{Liu2019DARTSDA} can not reflect the performance of its discretized sub-networks, resulting in sub-optimal search results.\footnote{The performance of the original DARTS will be presented as ablation studeis in Table \ref{tab:results_ablation_search_algorithms}, supporting our claims.} This result motivates us to propose a simple-yet-effective modification to the DARTS-style architecture search. Instead of relying on the architecture weights' values to select the best operation, we propose directly evaluating the operation's superiority by its influence on the hyper-network's performances. Since our method mimics the process of conducting ablation studies of a certain operation on a node from the hyper-network, we refer to our method as the ablation-based differentiable architecture search (AB-DARTS).

\subsection{Calculating the superiority of each operation}

We first introduce the core of our AB-DARTS method: the calculation of each operation's superiority score, defined as how much it affects the performance of the hyper-network. Denote the complete hyper-network as $M$. Hyper-network $M$ is trained till convergence on the training set. We now consider a modified hyper-network obtained by masking out an operation $o_{i, j}$ on $\text{Node}_{i}$ while keeping all other operations on the node. This new hyper-network is denoted as $M_{\backslash o_{i, j}}$. We evaluate the two versions of hyper-networks on the validation data $D_{val}$. Denote the metric score as a function of a model $M$, $S(M)$, with the validation data fixed. Then the contribution score of operation $o_{i, j}$ is given by:
\begin{equation}
\text{CS}(o_{i, j}) = S(M) - S(M_{\backslash o_{i, j}}).
\end{equation}
Note that in the above equation, $S(M)$ can be treated as a constant term. Thus the above equation can be simplified to $\text{CS}(o_{i, j}) = - S(M_{\backslash o_{i, j}})$. The operation that results in the most significant drop upon masking in the hyper-network's validation metric will be considered the most critical operation on that edge. In the experiments, we set $S()$ as the negative of the cross-entropy (CE) loss since the widely applied metrics like accuracy may not vary if the hyper-network only masks out a single operation. 


\subsection{The complete process of AB-DARTS}


We now describe the whole process of our AB-DARTS method. Our AB-DARTS method requires the hyper-network to be trained for $K_1 > 0$ epochs until convergence on the train set. Note that different from \cite{liu2018darts}, we freeze the architectural parameters and train only the model parameters on the train set $D_{train}$. No bi-level optimization is required, thus saving training time costs. Then, we traverse over all the operations on every node. For each node $\text{Node}_{i}$ of the hyper-network, we evaluate the superiority of each operation $o_{i, j}$ on the development set $D_{val}$. Then we select the operation $o_{i}^{*}$ that receives the highest superiority score for discretizing this node, that is, keeping this operation and dropping all other operations on this node. After the discretization of a node, we tune the altered hyper-network for $K_2 > 0$ epochs to make the remaining hyper-network recover the lost performance. The above steps are repeated until all the nodes are discretized for the hyper-network, and we obtain the sub-network's architecture. Then the sub-network $M_{S}$ is randomly initialized and trained on the $D_{train}$ for $K_3 > 0$ epochs. Formally, we summarize the above process in Algorithm \ref{algo:ab_darts}.

\begin{algorithm}[!ht]
\DontPrintSemicolon
  
\KwInput{Hyper-network $M$ with $N$ nodes; }
\KwOutput{Sub-network $M_{S}$, with the set of selected operations $\{ o^{*}_{i} \}_{i=1}^{N} $ }
\KwData{Training set $D_{train}$, validation data $D_{val}$ }
    

Train the hyper-network $M$ on the training set $D_{train}$ for $K_1$ epochs; 

\For{node index $1 \leq i \leq N$ }
{
   \For{operation $o_{i, j}$ on node $\text{Node}_{i}$ }
   {
    Calculate the superiority score $\text{CS}(o_{i, j})$ on $D_{val}$;
   }
   Select the best operation $o_{i}^{*} \leftarrow \arg \max_{j} \text{CS}(o_{i, j})$;
   
   Discretize $\text{Node}_{i}$ of hyper-network $M$ by only keeping $o_{i}^{*}$;
   
   Further train the hyper-network $M$ on $D_{train}$ for $K_2$ epochs; 
 
}

Train the sub-network $M_{S}$ on the training set $D_{train}$ for $K_3$ epochs; 

\caption{Ablation-based differentiable architecture search}
\label{algo:ab_darts}
\end{algorithm}

\section{Experiments}

\subsection{Baselines} 

We compare our AutoFormer-TS method with the three groups of SOTA time series models on the long-horizon forecasting tasks: (a) deep learning based models, including DLinear \cite{zeng2023transformers} and TimesNet \cite{wu2022timesnet}; (b) Transformer-based models, including FedFormer \cite{zhou2022fedformer}, Autoformer \cite{wu2021autoformer}, Informer \cite{zhou2021informer}, Reformer \cite{kitaev2020reformer}, iTransformer \cite{liu2023itransformer} and PatchTST \cite{nie2022time}; (c) Time-series models based on pretrained models, either pretrained on the large-scale time series corpus or the other modality. These types of methods include MOMENT \cite{goswami2024moment}, Time-LLM \cite{jin2023time}, GPT4TS \cite{zhou2023one}. On the short-term forecasting tasks, we further compare our model with N-HiTS \cite{challu2023nhits} and N-BEATS \cite{oreshkin2019n}.

\subsection{Datasets and evaluation metrics}

For long-term time series forecasting, we assess our Time-LlaMA framework on the following datasets, in accordance with \cite{wu2022timesnet}: ETTh1, ETTh2, ETTm1, ETTm2, Weather, Electricity (ECL), Traffic and ILI. These tasks have been extensively adopted for benchmarking long-term forecasting models. The input time series length $T_{L}$ is set as 512, and we use four different prediction horizons $T_{P} \in \{96, 192, 336, 720\}$. The evaluation metrics utilized are the mean square error (MSE) and the mean absolute error (MAE). 

For short-term time series forecasting, we employ the M4 benchmark \cite{makridakis2018m4}. This benchmark contains a
collection of marketing data in different sampling frequencies. The prediction horizons are relatively small and in \{6, 48\}, and The input lengths are twice as prediction horizons. The evaluation metrics for this benchmark include the symmetric mean absolute percentage error (SMAPE), the mean scaled absolute error (MSAE), and the overall weighted average (OWA). 

The introduction to the datasets and the evaluation metrics are presented in Appendix \ref{sec:appendix_datasets_and_metrics}.

\begin{table*}
\centering

\caption{\label{tab:results_main_long_term} Results for the long-term forecasting tasks. The prediction horizon $T_{P}$ is one of $\{24, 36, 48, 60\}$ for ILI and one of \{96, 192, 336, 720\} for the others. Lower value indicates better performance. \textbf{Bold} values represent the best score, while \underline{Underlined} means the second best score.} 

\resizebox{1.0\textwidth}{!}{
\renewcommand\arraystretch{1.35}
\begin{tabular}{c|cc|cc|cc|cc|cc|cc|cc|cc}
\toprule
\multirow{2}*{ \textbf{Task}  }    &   \multicolumn{2}{c}{ETTh1}    &   \multicolumn{2}{c}{ETTh2}   &   \multicolumn{2}{c}{ETTm1}    &  \multicolumn{2}{c}{ETTm2}    &    \multicolumn{2}{c}{Weather}    &    \multicolumn{2}{c}{ECL}   &    \multicolumn{2}{c}{Traffic}  &    \multicolumn{2}{c}{ILI}       \\

& MSE   &  MAE    &  MSE   &  MAE     &  MSE   &  MAE     &  MSE   &  MAE     &  MSE   &  MAE    &  MSE   &  MAE  &  MSE   &  MAE  &  MSE   &  MAE   \\

\midrule

DLinear   &   0.422 & 0.437 & 0.431 & 0.446 & 0.357 & \underline{0.378} & 0.267 & 0.333 & 0.248 & 0.300 & 0.166 & 0.263 & 0.433 & 0.295 & 2.169 & 1.041\\
TimesNet    &   0.458 & 0.450 & 0.414 & 0.427 & 0.400 & 0.406 & 0.291 & 0.333 & 0.259 & 0.287 & 0.192 & 0.295 & 0.620 & 0.336 & 2.139 & 0.931    \\

Autoformer    &   0.496 & 0.487     &   0.450 & 0.459   &   0.588 & 0.517   &   0.327 & 0.371   &   0.338 & 0.382   &   0.227 & 0.338   &   0.628 & 0.379   &   3.006 & 1.161  \\
Informer   &  1.040 & 0.795  &   4.431 & 1.729   &  0.961 & 0.734  &  1.410 & 0.810  &  0.634 & 0.548  &  0.311 & 0.397  &   0.764 & 0.416  &   5.137 & 1.544 \\
Reformer    &  1.029 & 0.805 & 6.736 & 2.191 & 0.799 & 0.671 & 1.479 & 0.915 & 0.803 & 0.656 & 0.338 & 0.422 & 0.741 & 0.422 & 4.724 & 1.445    \\
FedFormer   &   0.440 & 0.460   &  0.437 & 0.449   &  0.448 & 0.452   & 0.305 & 0.349   &  0.309 & 0.360   &  0.214 & 0.327   & 
  0.610 & 0.376   &   2.847 & 1.144  \\

iTransformer    &   0.440 & 0.460    &  0.437 & 0.449    &  0.448 & 0.452    &  0.305 & 0.349    &  0.309 & 0.360    &  0.214 & 0.327    &  0.610 & 0.376    &  2.847 & 1.144  \\
PatchTST    &   \underline{0.413} & \underline{0.430} & \underline{0.330} & \underline{0.379} & 0.351 & 0.380 & \underline{0.255} & \underline{0.315} & \textbf{0.225} & 0.264 & \underline{0.161} & \underline{0.252} & \underline{0.390} & \underline{0.263} & \underline{1.443} & \underline{0.797}  \\

\midrule
MOMENT   &   0.418   &  0.436   &   0.352   &    0.394   &  0.344  &   0.379   &  0.278   &   0.321     &  0.228   &  0.269    &   0.171   &    0.272      &  0.410    &   0.297    &    1.952    &  1.107    \\
Time-LLM   &  0.428  &   0.433   &    0.344  &   0.393    &  \underline{0.339}   &   0.382   &    0.271   &   0.319    &  \underline{0.225}   &   \underline{0.257}    &  
  0.168    &    0.262   &   0.408   &    0.286   &   1.835   &   0.906  \\

GPT4TS   &   0.465 & 0.455  &  0.381 & 0.412  &   0.388 & 0.403  & 0.284 & 0.339 &  0.237 & 0.270 &   0.167 & 0.263 &  0.414 &  0.294 &   1.925 & 0.903  \\

\midrule
AutoFormer-TS (ours)   &    \textbf{0.407} & \textbf{0.424}   &   \textbf{0.327} & \textbf{0.374}   &   \textbf{0.329} & \textbf{0.372}   &   \textbf{0.251} & \textbf{0.313}   &   0.225 & \textbf{0.254}   &  \textbf{0.158} & \textbf{0.252}   &   \textbf{0.388} & \textbf{0.264}   &   \textbf{1.435} & \textbf{0.801}   \\

\bottomrule
\end{tabular}}

\end{table*}

\subsection{Experimental setups}

\noindent \textbf{Devices} \quad We run all our experiments on NVIDIA 4090ti (24GB) GPUs.

\noindent \textbf{Hyper-parameters for the architecture search stage} \quad Our AB-DARTS method in our AutoFormer-TS framework divides the workflow into two stages, the architecture search stage and architecture training stage. During the architecture search stage, the hyper-network is initialized with the weighted operations and all the architectural parameters are uniformly initialized. The hyper-network has three Transformer blocks, and the hidden size is 256. For time series tokenization, we utilize the patching strategy introduced in \cite{nie2022time}, with patch length 16 and stride 8. The time series patches are projected to the model dimension via a linear layer. The forecasting head follows \cite{nie2022time}, that is, all the time series token vectors are concatenated and fed into a linear layer to obtain the predicted time series values. 

For training the hyper-network, we use AdamW \cite{Loshchilov2019DecoupledWD} as the optimizer with a linear learning rate decay schedule and 6\% of the training steps for warm-up. The learning rate is set to 1e-4. The batch size is set according to the training set size, so that each training epoch contains around 256 to 512 optimization steps. During training, the architectural weights are frozen. The contribution score for each operation is calculated on the development set $D_{dev}$. For training epochs, we set $K_1 = 5$ and $K_2 = 1$. Once the hyper-network is fully discretized, we obtain the learned sub-network on the task at hand.

\noindent \textbf{Hyper-parameters for the learned architectures} \quad The hyper-parameters for the learned architectures that is related to the model sizes will be kept the same with the search stage. The number of Transformer blocks is 3, and different blocks may have different architectures, that is, different attention mechanisms, different activation functions, or different encoding operations.

\noindent \textbf{Hyper-parameters for the learned architectures' training and evaluation} \quad For training the learned architectures, we keep the training settings almost the same with the search stage. Except that during training at this stage, no contribution scores are needed. The number of epochs is $K = 50$. Early stopping with maximum patience 10 is set. The model is evaluated at the development set $D_{dev}$ every 100 training steps. If the patience exceeds 10, that is, the model's evaluation performance does not improve for the last ten evaluation run, then the model will stop training. And the checkpoint with the best evaluation performance will be used to make predictions on the test set.  

\noindent\textbf{Reproducibility} \quad We run the search stage once for each task. And the learned sub-network for each task will be run under five different random seeds and the mean performance on the test set will be reported.

\subsection{Main results}

\noindent \textbf{Results for long-term forecasting} \quad In Table \ref{tab:results_main_long_term}, we report the average score over four different prediction horizons for the long-horizon time series forecasting tasks. The experimental results demonstrate that our AutoFormer-TS method outperforms the baselines on most of the (task, prediction horizon) pairs. When compared to the previous state-of-the-art (SOTA) model PatchTST, Time-LlaMA can also achieves advantages. The comparisons against MOMENT, Time-LLM and GPT4TS are also meaningful. These three are very recent works on adapting large language models to the time-series forecasting tasks.

\noindent \textbf{Results for short-term forecasting} \quad To demonstrate that our method works in the short-term forecasting tasks, we utilize the M4 benchmark \cite{makridakis2018m4}. Table \ref{tab:results_main_short_term} reports the SMAPE, MSAE and OWA scores. Our experimental results demonstrate that our AutoFormer-TS method consistently surpasses all baselines when conducting short-term time series predictions.

\begin{table*}[tb!]
\centering
\caption{\label{tab:results_main_short_term} Results for the short-term time series forecasting task, M4. The forecasting horizons are in \{6, 48\}. Lower value indicates better performance. \textbf{Bold} values represent the best score, while \underline{Underlined} means the second best. } 

\resizebox{1.0\textwidth}{!}{
\renewcommand\arraystretch{1.3}
\begin{tabular}{cc|cccccccccc}
\toprule
\multicolumn{2}{c}{\textbf{Method}}   &   \textbf{AutoFormer-TS (ours)}    &   \textbf{GPT4TS}  &   \textbf{TIME-LLM}    &      \textbf{MOMENT}  &    \textbf{PatchTST}   &     \textbf{DLinear}    &   \textbf{TimesNet}    &   \textbf{FEDformer}    &   \textbf{N-HiTS}    &     \textbf{N-BEATS}    \\ 

\midrule

\multirow{3}*{ Yearly  }   &  SMAPE   &   \textbf{13.11}  &  15.11   &   13.62    & 13.72    & 13.68  &  16.96   &  15.37  & 14.02   & \underline{13.42}   & 13.48    \\
&   MSAE    &   \textbf{3.01}     & 3.56   &   3.06  & 3.07    & 3.12   & 4.28  & 3.55   & 3.04  & 3.06 &  \underline{3.04}  \\
&   OWA     &   \textbf{0.78}   & 0.91   &   \underline{0.79}  &  0.81  & 0.79  & 1.06   & 0.92   & 0.81    &   0.80   &   0.79  \\

\midrule

\multirow{3}*{ Quarterly  }   &  SMAPE   &   \textbf{10.06}  &   10.59   &   10.52  & 10.96    & 10.38  & 12.14  & 10.46  & 11.10  & \underline{10.19} & 10.56 \\
&   MSAE    &   \textbf{1.16}   & 1.25  &   \underline{1.18}  & 1.32    & 1.23  & 1.52  & 1.22 &  1.35  & 1.18 & 1.25 \\
&   OWA     &   \underline{0.89}   & 0.94   &   0.89   & 0.98   & 0.92  & 1.11  & 0.92 & 0.99  & \textbf{0.89} & 0.93 \\

\midrule

\multirow{3}*{ Monthly  }   &  SMAPE   &  \textbf{12.96}  & 13.26  &   13.10  & 13.92   & \underline{12.96}   & 13.51  & 13.51 & 14.40   & 13.06  & 13.09   \\
&   MSAE    &   \underline{0.99}   & 1.00   &   1.10  & 1.09   & 0.97   & 1.03   & 1.03  & 1.14  & 1.01 &  \textbf{0.99}  \\
&   OWA     &   \textbf{0.90}   & 0.93   &   0.94  & 0.99   &  \underline{0.91}  & 0.95    & 0.95  & 1.03  & 0.92 & 0.92  \\

\midrule

\multirow{3}*{ Others  }   &  SMAPE   &   \textbf{4.51}  & 6.12  &   4.93 & 6.30   & 4.95   & 6.70   & 6.91  & 7.15   &   \underline{4.71} & 6.59  \\
&   MSAE    &   \textbf{2.81}  & 4.11   &   \underline{2.95}  & 4.06  & 3.34  & 4.95  & 4.50  & 4.04 & 3.05 & 4.43 \\
&   OWA     &   \textbf{0.94}  & 1.25   &   \underline{0.96}  & 1.30   & 1.04  & 1.48    & 1.43  & 1.38   & 0.97 & 1.39   \\

\midrule

\multirow{3}*{ Average  }   &  SMAPE   &  \textbf{11.69} & 12.69   &   \underline{11.92}  & 12.78  & 12.05   & 13.63  & 12.88   & 13.16  & 12.035 & 12.25 
 \\
&   MSAE    &   \textbf{1.54}   & 1.808   &   \underline{1.57}  & 1.75   & 1.62  & 2.09   & 1.83   & 1.77  & 1.62 & 1.69 \\
&   OWA     &   \textbf{0.81}   & 0.94    &   \underline{0.87}  & 0.93  & 0.86  & 1.05  & 0.95  & 0.94  & 0.87 & 0.89 \\

\bottomrule
\end{tabular}}

\end{table*}

\begin{figure*}[tb!]
\centering
\includegraphics[width=0.86\textwidth]{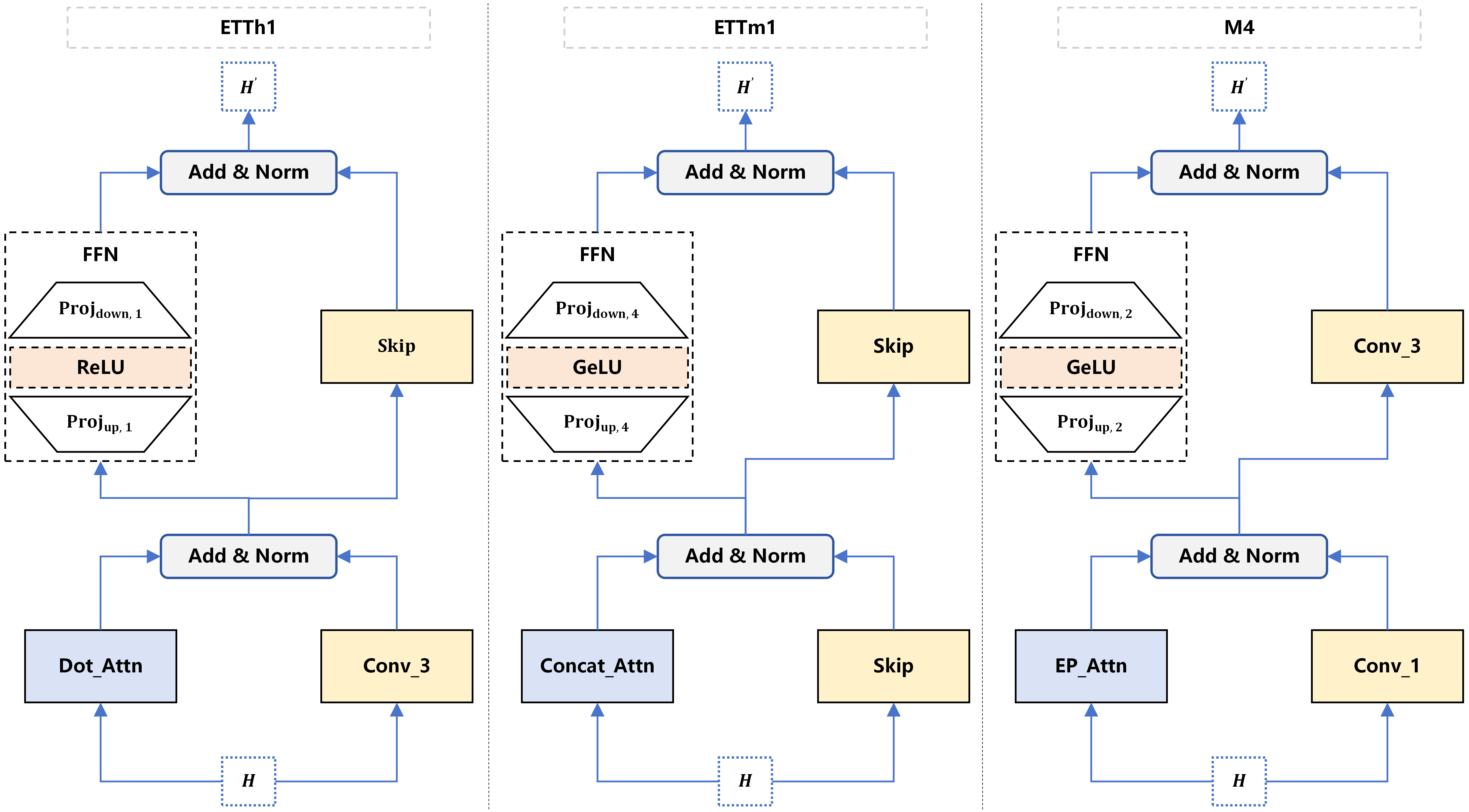}
\caption{The learned architectures on the ETTh1, ETTm1 and M4 tasks. }
\label{fig:learned_archs}
\end{figure*}

\noindent \textbf{Visualization of the learned architectures} \quad We present the third Transformer layers of the learned architectures on the ETTh1, ETTm1 and M4 tasks in Figure \ref{fig:learned_archs}. The whole detailed architectures on each of the tasks are presented in Table \ref{tab:learned_architectures} in Appendix \ref{sec:appendix_learned_architectures}. From the learned architectures, we can observe that:
\begin{itemize}

\item We observe task specificity for the best architectures found on different tasks are different, emphasizing the importance of task specificity. Note that task specificity is needed to obtain the SOTA performance on each task, but we will also show that the architectures have transferability to a certain degree on Table \ref{tab:results_transfer}.

\item We notice that the task with smaller dataset size prefers a more light-weighted architecture. For example, the ETTh1's model set the dimension multiplication factor $k$ to be $0.5$ at the second layer, and $1$ for the first and third layers.

\item We observe that in many of the tasks, the convolution operations act as the encoding operations accompanying the self-attention module and FFN modules. These design patterns are also observed in \cite{zhu2021autonlu,so2019evolved,jiang2020convbert}. Intuitively, convolution operations extract local features similar to n-gram, which complement long-term dependency features captured by self-attention. 

\end{itemize}

\subsection{Ablation studies and further analysis}

\noindent\textbf{Ablations on the search algorithm} \quad Note that the main experiments (Table \ref{tab:results_main_long_term} and \ref{tab:results_main_short_term}) employ our AB-DARTS method (Section \ref{sec:search_algorithm}) to conduct search on our search space. In order to demonstrate the effectiveness of our method, we now substitute the search algorithm by a series of baseline neural architecture search algorithms: (a) DARTS \cite{liu2018darts}; (b) Stable-DARTS \cite{bi2019stabilizing}; (c) Gold-NAS \cite{bi2020gold}; (d) ENAS \cite{pham2018efficient}. The experiments are conducted on the ETTh1, ETTm1 and M4-Yearly tasks, and the results are reported in Table \ref{tab:results_ablation_search_algorithms}. The experimental results show that the AB-DARTS result in the best performance, demonstrating its effectiveness in selecting appropriate operations for the task at hand. The effectiveness of our AutoFormer-TS framework comes from the superiority scoring method for the operations during architecture search.

\begin{table}
\centering
\caption{\label{tab:results_ablation_search_algorithms} Results for the ablation studies on the architecture search algorithms. } 
\resizebox{0.46\textwidth}{!}{
\renewcommand\arraystretch{1.3}
\begin{tabular}{c|c|c|cc}
\toprule
\multirow{2}*{ \textbf{Task}  }    &   ETTh1    &    ETTm1    &      \multicolumn{2}{c}{M4}      \\

& MSE    &  MSE    &    SMAPE   &   MSAE   \\

\midrule

A-DARTS (ours)   &    \textbf{0.407} &     \textbf{0.329}    &    \textbf{13.11} &    \textbf{3.01}  \\

\midrule
DARTS   &   0.412   &  0.346     &    13.35    &  3.21   \\
Stable-DARTS   &    0.411     &   0.341     &  13.29 &  3.19       \\ 
Gold-NAS       &   0.415   &   0.343     &  13.15   &  3.17   \\
ENAS   &    0.413    &  0.351   &     13.47  &   3.24    \\

\bottomrule
\end{tabular}}
\end{table}

\noindent\textbf{Ablation studies on the search space} \quad We now conduct an ablation study of our search space by reducing our search space $\mathcal{S}$ to a singleton containing the vanilla Transformer architecture step-by-step: (a) reduce the search space by restricting that the Transformer architecture must be identical across different blocks (denoted as search space $\mathcal{S}_1$). This type of search space is referred to as the micro search space \cite{liu2018darts}, whereas the one in our main experiments is referred to as the macro search space. Intuitively, the macro search space is much larger than the micro one by allowing more flexible architectural designs. (b) reduce the activation's search space by only keeping the \textbf{ReLU} activation for $g())$ (denoted as search space $\mathcal{S}_2$); (c) further reduce the encoding operations' search space to only include \textbf{skip} ($\mathcal{S}_3$); (c) further restrict the dimension multiplication factor $k$ in the FFN module to $k=4$ ($\mathcal{S}_4$). Note that $\mathcal{S}_4$ contains only the vanilla Transformer in PatchTST \cite{nie2022time}. Table \ref{tab:results_ablation_search_spaces} reports the results on different search spaces, showing that that dropping any components of the whole search space results in performance losses. The results demonstrate that each components of the search space is necessary and beneficial.

\begin{table}
\centering
\caption{\label{tab:results_ablation_search_spaces} Results for the ablation studies on the search spaces. } 
\resizebox{0.42\textwidth}{!}{
\renewcommand\arraystretch{1.3}
\begin{tabular}{c|c|c|cc}
\toprule
\multirow{2}*{ \textbf{Task}  }    &   ETTh1    &   ETTm1    &    \multicolumn{2}{c}{M4-Yearly}      \\

& MSE       &  MSE         &    SMAPE   &   MSAE   \\

\midrule

$\mathcal{S}$   &    \textbf{0.407} &     \textbf{0.329}    &   \textbf{13.11} &    \textbf{3.01}   \\

\midrule

$\mathcal{S}_1$   &    0.409   &   0.331   &   13.16   &  3.05   \\
$\mathcal{S}_2$   &   0.410      &  0.342     &  13.45   &  3.08 \\
$\mathcal{S}_3$   &  0.410    &   0.345     &   13.57  &  3.10 \\
$\mathcal{S}_4$ (PatchTST)  &   0.413   &   0.351   &   13.68   &  3.12   \\

\bottomrule
\end{tabular}}
\end{table}


\noindent\textbf{Transferability across tasks} \quad Note that the main experiments (Table \ref{tab:results_main_long_term} and \ref{tab:results_main_short_term}) demonstrate task specificity under the AutoFormer-TS framework, that is, the learned sub-networks are different across tasks, and achieve SOTA performances. Now we will demonstrate that the learned sub-networks can be transferred to other tasks and achieve reasonable performances. The transferability experiments are conducted on ETTh1, ETTm1 and M4-Yearly, and the results are presented in Table \ref{tab:results_transfer}. In Table \ref{tab:results_transfer}, each row presents the learned models on different tasks, and each column presents the target task. From Table \ref{tab:results_transfer}, the following observations can be made:
\begin{itemize}
\item The best performance are obtained by learning the model on the task at hand, and the transferred model from other tasks performs less well. 

\item Note that on the two long-horizon forecasting tasks, the transferred models obtain better performance than PatchTST, showing a certain degree of transferability. 

\item transferability between the long-horizon and short-horizon forecasting tasks are less well. The transferred models performs worse than PatchTST. 
\end{itemize}

\begin{table}
\centering
\caption{\label{tab:results_transfer} Results for transferring the learned architectures from one task to another. } 
\resizebox{0.48\textwidth}{!}{
\renewcommand\arraystretch{1.1}
\begin{tabular}{c|c|c|cc}
\toprule
\multirow{2}*{ \textbf{Model}  }    &   ETTh1    &   ETTm1    &    \multicolumn{2}{c}{M4-Yearly}      \\

& MSE       &  MSE         &    SMAPE   &   MSAE   \\

\midrule

Learned model on ETTh1   &   \textbf{0.407}   &   0.335   &   13.85   & 
 3.18       \\
Learned model on ETTm1   &   0.411   &   \textbf{0.321}   &   13.85   & 
 3.18       \\

 Learned model on M4-Yearly   &   0.418  &   0.367  &   \textbf{13.11} &    \textbf{3.01}   \\

 PatchTST    &   0.413   &   0.351   &   13.68   &  3.12  \\

\bottomrule
\end{tabular}}
\end{table}

\noindent\textbf{On the efficiency of the AutoFormer workflow} \quad  We use the ETTh1 task to demonstrate the search efficiency. Running the ETTh1 task with PatchTST takes 10.2 min. For searching, DARTS takes 30.6 min for searching new architectures and another 10.5 min for running the obtained architecture. Since AB-DARTS does not require bi-level optimization, it requires 11.5 min for running training steps and 9.4 min for calculating operations' contribution scores. And the re-training stage takes 9.97 min. Our method consumes around three times the training time of PatchTST, which is affordable compared to manually designing different architectures and running numerous evaluations.

\section{Conclusion}

In this work, we propose the AutoFormer-TS framework, which enhances the performance of Transformer on time series forecasting tasks via searching for novel architectures. First, a novel and comprehensive search space is constructed, allowing for room of improvements via architectural design. Then we introduce a novel AB-DARTS method. This method improves upon existing DNAS approaches by better selecting the proper neural network operations. Extensive experiments proves that AutoFormer-TS consistently outperforms state-of-the-art baselines across various forecasting benchmarks. In addition, our framework is efficient since it does not require too much additional training time.


\bibliographystyle{ACM-Reference-Format}
\bibliography{sample-base}

\appendix

\section{Appendix: datasets and evaluation metrics}
\label{sec:appendix_datasets_and_metrics}

\subsection{Datasets}

We evaluate the long-term forecasting (ltf) performance on the well-established eight different benchmarks, including four ETT datasets  (including ETTh1, ETTh2, ETTm1, and ETTm2) from \cite{zhou2021informer}, Weather, Electricity, Traffic, and ILI from \cite{wu2021autoformer}. For short-term time series forecasting (STF), we employ the M4 benchmark \cite{makridakis2018m4}.

\noindent\textbf{ETT} The Electricity Transformer Temperature (ETT) is a crucial indicator in the electric power long-term deployment. This dataset consists of 2 years data from two separated counties in China. To explore the granularity on the Long sequence time-series forecasting (LSTF) problem, different subsets are created, {ETTh1, ETTh2} for 1-hour-level and ETTm1 for 15-minutes-level. Each data point consists of the target value ”oil temperature” and 6 power load features. The train/val/test is 12/4/4 months.

\noindent\textbf{ECL} Measurements of electric power consumption in one household with a one-minute sampling rate over a period of almost 4 years. Different electrical quantities and some sub-metering values are available.This archive contains 2075259 measurements gathered in a house located in Sceaux (7km of Paris, France) between December 2006 and November 2010 (47 months).

\noindent\textbf{Traffic} Traffic is a collection of hourly data from California Department of Transportation, which describes the road occupancy rates measured by different sensors on San Francisco Bay area freeways.

\noindent\textbf{Weather} Weather is recorded every 10 minutes for the 2020 whole year, which contains 21 meteorological indicators, such as air temperature, humidity, etc. 

\noindent\textbf{ILI} The influenza-like illness (ILI) dataset
contains records of patients experiencing severe influenza with complications. 

\noindent\textbf{M4} The M4 benchmark comprises 100K time series, amassed from various domains commonly present in business, financial, and economic forecasting. These time series have been partitioned into six distinctive datasets, each with varying sampling frequencies that range from yearly to hourly. These series are categorized into five different domains: demographic, micro, macro, industry, and finance.

The datasets' statistics are presented in Table \ref{tab:dataset_stats}.

\begin{table*}
\centering
\caption{Dataset statistics. The dimension indicates the number of time series (i.e., channels), and the dataset size is organized in (training, validation, testing).}
\resizebox{0.92\textwidth}{!}{
\renewcommand\arraystretch{1.1}
\begin{tabular}{@{}ll|c|c|c|c|c}
\hline
\textbf{Tasks} & \textbf{Dataset} & \textbf{Dim.} & \textbf{Series Length} & \textbf{Dataset Size} & \textbf{Frequency} & \textbf{Domain} \\ 
\hline
\multirow{7}{*}{Long-term Forecasting} & ETTm1 & 7 & \{96, 192, 336, 720\} & (34465, 11521, 11521) & 15 min & Temperature \\
& ETTm2 & 7 & \{96, 192, 336, 720\} & (34465, 11521, 11521) & 15 min & Temperature \\
& ETTh1 & 7 & \{96, 192, 336, 720\} & (8545, 2881, 2881) & 1 hour & Temperature \\
& ETTh2 & 7 & \{96, 192, 336, 720\} & (8545, 2881, 2881) & 1 hour & Temperature \\
& Electricity & 321 & \{96, 192, 336, 720\} & (18317, 2633, 5261) & 1 hour & Electricity \\
& Traffic & 862 & \{96, 192, 336, 720\} & (12185, 1757, 3509) & 1 hour & Transportation \\
& Weather & 21 & \{96, 192, 336, 720\} & (36792, 5271, 10540) & 10 min & Weather \\
& ILI & 7 & \{24, 36, 48, 60\} & (617, 74, 170) & 1 week & Illness \\ \hline

\multirow{6}{*}{Short-term Forecasting} & M4-Yearly & 1 & 6 & (23000, 0, 23000) & Yearly & Demographic \\
& M4-Quarterly & 1 & 8 & (24000, 0, 24000) & Quarterly & Finance \\
& M4-Monthly & 1 & 18 & (48000, 0, 48000) & Monthly & Industry \\
& M4-Weakly & 1 & 13 & (359, 0, 359) & Weakly & Macro \\
& M4-Daily & 1 & 14 & (4227, 0, 4227) & Daily & Micro \\
& M4-Hourly & 1 & 48 & (414, 0, 414) & Hourly & Other \\ 
\hline

\end{tabular}}

\label{tab:dataset_stats}
\end{table*}

\subsection{Evaulation metrics}

We now specify the evaluation metrics we used for comparing different models. We utilize the mean square error (MSE) and mean absolute error (MAE) for long-term forecasting. For the short-term forecasting task on M4 benchmark, we adopt the symmetric mean absolute percentage error (SMAPE), mean absolute scaled error (MASE), and overall weighted average (OWA), following \cite{oreshkin2019n}. The calculations of these metrics are as follows:
\begin{align}
\text{MSE} &= \frac{1}{H} \sum_{h=1}^{T} (\mathbf{Y}_h - \hat{\mathbf{Y}}_h)^2,  \\
\text{MAE} &= \frac{1}{H} \sum_{h=1}^{H} |\mathbf{Y}_h - \hat{\mathbf{Y}}_h|,  \\
\text{SMAPE} &= \frac{200}{H} \sum_{h=1}^{H} \frac{|\mathbf{Y}_h - \hat{\mathbf{Y}}_h|}{|\mathbf{Y}_h| + |\hat{\mathbf{Y}}_h|},  \\
\text{MAPE} &= \frac{100}{H} \sum_{h=1}^{H} \frac{|\mathbf{Y}_h - \hat{\mathbf{Y}}_h|}{|\mathbf{Y}_h|},  \\
\text{MASE} &= \frac{1}{H} \sum_{h=1}^{H} \frac{|\mathbf{Y}_h - \hat{\mathbf{Y}}_h|}{\frac{1}{H-s} \sum_{j=s+1}^{H} |\mathbf{Y}_j - \mathbf{Y}_{j-s}|},  \\
\text{OWA} &= \frac{1}{2} \left[ \frac{\text{SMAPE}}{\text{SMAPE}_{\text{Naive}}} + \frac{\text{MASE}}{\text{MASE}_{\text{Naive}}} \right],  \\
\end{align}
where $ s $ is the periodicity of the time series data. $ H $ denotes the number of data points (i.e., prediction horizon in our cases). $ \mathbf{Y}_h $ and $ \hat{\mathbf{Y}}_h $ are the $ h $-th ground truth and prediction where $ h \in \{1, \cdots, H\} $.

\section{Appendix: the learned architectures on different tasks}
\label{sec:appendix_learned_architectures}

The Table \ref{tab:learned_architectures} presents the learned architectures via our AutoFormer-TS framework.

\begin{table*}
\centering
\caption{\label{tab:learned_architectures}The learned architectures on different tasks.}
\resizebox{0.98\textwidth}{!}{
\begin{tabular}{lccccc}
\hline \bf Architecture  & \bf ETTh1   &    \bf ETTh2   &  \bf ETTm1   &   \bf ETTm2   &  \bf   M4-yearly   \\ 
\hline

Layer 1: Encoding operation at self-attention  &   Conv\_3   &  Conv\_3   &   Conv\_3   &   Skip   &  Conv\_5   \\
Layer 1: Encoding operation at FFN  &    Skip  &   Skip  &    Skip  &   
  Conv\_1   &  Skip   \\
Layer 1: Attention function   &   Concat\_Attn &  Concat\_Attn  &   EP\_Attn &   Minus\_Attn   &  Dot\_Attn  \\
Layer 1: Activation function    &  Leaky\_ReLU    &  Leaky\_ReLU   &  ELU   &  GeLU   &    ReLU  \\
Layer 1: Dimension multiplication factor $k$  &   1 &   2   &  4 &   2  &  1  \\

\hline

Layer 2: Encoding operation at self-attention  &  Skip   &  Conv\_3   &   Conv\_1   &   Conv\_1   &  Conv\_3   \\
Layer 2: Encoding operation at FFN  &   Skip   &  Conv\_1  &  Skip   &  Skip  &  Conv\_3    \\
Layer 2: Attention function   &  Minus\_Attn   &   Minus\_Attn   &   Bilinear\_Attn   &  Dot\_Attn   &  Dot\_Attn    \\
Layer 2: Activation function    &   SWISH    &  GeLU   &  SWISH   &  GeLU   &    Leaky\_ReLU  \\
Layer 2: Dimension multiplication factor $k$  &  0.5 &   1   &  4 &   2  &  0.5   \\

\hline

Layer 3: Encoding operation at self-attention  &   Conv\_3   &    Conv\_1  &   Skip   &   Skip   &   Conv\_1  \\
Layer 3: Encoding operation at FFN  &   Skip  &     Conv\_5   &   Conv\_3   &  Skip    &   Conv\_3  \\
Layer 3: Attention function   &   Dot\_Attn    &      Concat\_Attn    &       EP\_attn   &     EP\_attn   &     Dot\_Attn  \\
Layer 3: Activation function    &   ReLU    &  GeLU   &  GeLU   &   Leaky\_ReLU  &  GeLU  \\
Layer 3: Dimension multiplication factor $k$  &   1   &  2  &  4  &  1 &  2   \\

\hline
\end{tabular}}

\end{table*}

\end{CJK*}

\end{document}